\pdfoutput=1
\documentclass[11pt,a4paper]{article}
\usepackage[hyperref]{naaclhlt2019}
\usepackage{times}
\usepackage{latexsym}
\usepackage{amsmath}
\usepackage{graphicx}
\usepackage{color}
\usepackage{caption}
\usepackage{xspace}
\usepackage{booktabs}
\usepackage{float}

\usepackage{hyperref}
\hypersetup{
    linkcolor=black,
    filecolor=black,
    linkcolor=black,
    urlcolor=black}

\newcommand{\xhdr}[1]{{\noindent\bfseries #1.}}
\setcitestyle{nosort}

\aclfinalcopy %
\def\aclpaperid{1771} %

\newcommand{\cut}[1]{}

\DeclareMathOperator{\argmax}{arg\,max}

\newcommand{\naone}{N$_1$\xspace}
\newcommand{\natwo}{N$_2$\xspace}
\newcommand{\nai}{N$_i$\xspace}
\newcommand{\qnone}{qn$_1$\xspace}
\newcommand{\qntwo}{qn$_2$\xspace}
\newcommand{\asq}{\texttt{ASQ}\xspace}
\newcommand{\storycloze}{story cloze test\xspace}

\title{Asking the Right Question: \\ Inferring Advice-Seeking Intentions from Personal Narratives}

\author{Liye Fu \\
  Cornell University \\
  {\tt liye@cs.cornell.edu} \\ \And 
  Jonathan P. Chang \hspace{0.3in} \\ 
  Cornell University \hspace{0.3in} \\ 
  {\tt jpc362@cornell.edu \hspace{0.1in}} \\ \And 
  Cristian Danescu-Niculescu-Mizil \\
  Cornell University \\ 
  {\tt cristian@cs.cornell.edu}}

\date{}

\begin{document}

\maketitle
\begin{abstract}

People often share personal narratives in order to seek advice from others. To properly infer the narrator's intention, one needs to apply a certain degree of common sense and social intuition. To test the capabilities of NLP systems to recover such intuition, we introduce the new task of inferring what is the advice-seeking goal behind a personal narrative. We formulate this as a cloze test, where the goal is to identify which of two advice-seeking questions was removed from a given narrative.

The main challenge in constructing this task is finding pairs of semantically plausible advice-seeking questions for given narratives. To address this challenge, we devise a method that exploits commonalities in experiences people share online to automatically extract pairs of questions that are appropriate candidates for the cloze task.  This results in a dataset of over 20,000 personal narratives, each matched with a pair of related advice-seeking questions: one actually intended by the narrator, and the other one not. The dataset covers a very broad array of human experiences, from dating, to career options, to stolen iPads. 
We use human annotation to determine the degree to which the task relies on common sense and social intuition in addition to a semantic understanding of the narrative. 
By introducing several baselines for this new task we demonstrate its feasibility and identify avenues for better modeling the intention of the narrator.

\end{abstract}

\section{Introduction}
\label{sec:intro}

\begin{quote}
\small
\hspace{-0.2cm} ``Computers are useless.\\
\hspace{-0.2cm} They can only give you answers.'' - Pablo Picasso\\
\end{quote}
\vspace{-0.3cm}

\begin{figure}[ht!]

  \begin{tabular}{ p{7.5cm} }
      \toprule
    {\bf Personal narrative}: I am generally a person who needs a lot of sleep, but today I was not able to sleep more than 6 hours and I am extremely tired. My eyes hurt and two hours later I have programming [lesson]
     so I have to be alert. I've already drunk a cup of coffee and although I rarely drink coffee, it had no effect on me. I am not at home so I have limited possibilities as for food. I don't want to do anything too unhealthy such as drinking 10 cups of coffee, tho I may consider drinking another one. \\
    \midrule
    {\bf Which advice-seeking question is more likely to have been asked by the narrator:} \\
    {\bf Q1}: Is it even possible to be addicted to coffee?\\
    {\bf Q2}: How can I energize myself?\\
    \bottomrule
    \end{tabular}
\caption{An abbreviated instance from the \asq dataset. A personal narrative is matched with two plausible advice-seeking questions, only one of which was actually asked by the narrator when sharing the story.}
\label{fig:introexample}
\end{figure}

\begin{table*}[ht!]

\vspace{0.1in}
  \centering
  \begin{tabular}{p{0.5cm} p{5cm} p{7cm}} 
  \toprule
  & {\bf Task} &  {\bf Desired output} \\
 \midrule
A & Question generation & What do I need to do in 2 hours? \\

 & Reading comprehension & \\
 \addlinespace[3pt]

& Summarization  & I must go for a lesson after getting little sleep. \\

\midrule

B & Ending generation & Lastly, I tried an energizing drink. \\

 &  Narrative chains, story cloze \\

\midrule

C & Event2Mind &  to learn to code, to be educated \\
 & Desire fulfillment & \\
\bottomrule
\addlinespace[3pt]

D & {\bf Our task} & How can I energize myself? \\
  \bottomrule
\end{tabular}
\caption{Contrast with desired outputs in related narrative understanding tasks that focus on the within-story (intradiegetic) aspects of narrative understanding. Tasks are grouped according to the categories discussed in Section \ref{sec:task}. We assumed the second sentence (``My eyes hurt and two hours later I have programming lesson so I have to be alert.'') to be the answer span for question generation, and the input for Event2Mind (which operates at sentence level). 
}

\label{tab:tasks}
\end{table*}

People often share their personal experiences to elicit advice from others.
These personal narratives provide the necessary context for properly understanding the informational goals of the narrators.  
Endowing automated systems with the capability to infer these advice-seeking intentions could support 
personalized assistance and more 
empathetic
human-computer interaction.

As humans, to properly distill the narrator's intention from the events and situations they describe, we need to 
apply a certain degree of social intuition \cite{conzelmann_social_2012,conzelmann_new_2013,baumgarten_cue_2015,kehler_evaluating_2017}.
As an example, consider the goals of a narrator sharing the personal story in Figure~\ref{fig:introexample}.   
We are presented with a wealth of information about the narrator's general sleep patterns, about a particular sleep deprivation situation and its physiological effects, about an upcoming lesson, about coffee intake, its effects, and potential health impacts, and about the current location of the narrator and its impact on food supply.
Taking these facts separately, we can imagine providing advice on how to get more sleep, on whether to postpone the lesson, on how to get food delivered, or on the risks of caffeine intake.   
However, given how the narrative is constructed, we can intuit that the more likely goal of the narrator is to get advice on how to overcome the effects of sleep deprivation so that they can be alert for the upcoming programming %
lesson.

Importantly, the primary goal of our proposed task is not to understand details about the narrator's actions in the story (``Why is the narrator tired?'', ``When do they need to go to the lesson?''), but to infer the reason why the narrator is sharing this story (i.e., ``To get advice on how to stay alert in the next few hours.'').
That is, we are not concerned with the \emph{intradiegetic} aspects of the narrative, but with the \emph{extradiegetic} intention of the narrator in sharing the story.  
While an understanding of the former is likely necessary for the latter, it is often not sufficient.

In this work, we introduce a task and a large dataset to evaluate the capabilities of automated systems to infer the narrator's (extradiegetic) intention in constructing and sharing an advice-seeking personal story.  
This complements existing narrative understanding tasks which focus on testing 
 semantic understanding of events, actors and their (intradiegetic) intentions within the narrative itself.  
Table \ref{tab:tasks} contrasts the goals of these existing narrative understanding tasks with that of inferring a narrator's advice-seeking intention, in the context of our introductory example.

Formally, we implement the task as a binary choice
 cloze test, where the goal is to identify which of two candidate advice-seeking questions was actually asked by the narrator of a given personal narrative. Beyond collecting a large and diverse set of realistic personal stories that contain an advice-seeking question, the main challenge in constructing this task is finding a plausible alternative advice-seeking question for each given narrative.  
To address this challenge, we develop a methodology for identifying such questions by exploiting both the commonalities in experiences people share online and the diversity of possible advice-seeking intentions that can be tied to similar experiences.

By applying our methodology to a large collection of online personal narratives, we construct a dataset of over 20,000 cloze test instances, covering a very broad spectrum of realistic advice-seeking situations.%
\footnote{The dataset is available at \url{https://github.com/CornellNLP/ASQ}.} Each instance contains a narrative that is matched with two advice-seeking questions, one of which is actually asked by the narrator  (Q2 in our introductory example), and the other 
semantically related to the narrative (Q1). 

We use human annotations to judge the relative difficulty of different subsets of the test instances
and the type of reasoning necessary to solve them. 
We find that more than half of the instances contain pairs of questions that are not only semantically related to the narratives but also do not contain any explicit factual mismatches with the stories.  
These are thus unsolvable by pure logical reasoning and require some degree of common sense or social intuition.  
And indeed, simple baseline approaches perform worse on these types of instances, highlighting the need for more direct modeling of the intention of the narrator.

To summarize, in this work we:
\begin{itemize}
    \item formulate the task of inferring advice-seeking intents from personal narratives (Section~\ref{sec:data});
    \item develop a methodology to construct a large dataset of personal narratives matched with plausible options for \textbf{a}dvice-\textbf{s}eeking \textbf{q}uestions (the \asq dataset) to be used for this task (Section~\ref{sec:method});
    \item show the task is viable and evaluate the relative difficulty of its items (Sections~\ref{sec:eval} \&~\ref{sec:baseline}).
\end{itemize}

We end by discussing the practical implications of endowing systems with the capability to infer advice-seeking intentions and use our results to identify avenues for developing better models.

\section{Task formulation}
\label{sec:data}

To evaluate the capability of automated systems to infer advice-seeking intentions, we formulate a cloze-style binary choice test where the system is presented with a personal narrative and is required
  to choose between two plausible candidate questions: one actually asked by the narrator and the other one not (as exemplified in Figure \ref{fig:introexample}).

We motivate the task 
 by contrasting it with 
  other (narrative) understanding tasks (Section \ref{sec:task}), and provide the rationale for this particular formulation by discussing its advantages (Section \ref{sec:formulation}).

\subsection{Related narrative understanding tasks}
\label{sec:task}

There are many tasks involving reading comprehension in general, and story understanding in particular. Given a narrative, there are a few broad categories of questions that may be asked to test different 
types and degrees of understanding.  
Table \ref{tab:tasks} follows directly from the discussion below, by contrasting the goal of our task with those of (intradiegetic) narrative understanding tasks in the context of our introductory example.

\noindent\textbf{A: What happened in the story?} The most direct approach to test story understanding is to check whether the reader could comprehend the events and actions that occur within the story. 
This requires 
 semantic understanding, but nothing more. 
 This type of task can be set up in various forms, as the system can be asked to summarize 
  the story 
   (summarization, see \citet{nenkova_automatic_2011,allahyari_text_2017} for surveys), generate a question that is answerable from the text (question generation \cite{du_learning_2017}), or answer a question for which the information can be retrieved
       or reasoned 
    directly from the story (reading comprehension, see \citet{chen_neural_2018} for a survey; notable datasets include MCTest \cite{richardson_mctest_2013} and NarrativeQA \cite{kocisky_narrativeqa_2018}).

\noindent\textbf{B: What might happen next?} While reading the story, people not only grasp and process the events that 
already occurred but also have some intuition of its likely trajectory.
Related tasks include the narrative cloze task \cite{chambers_unsupervised_2008}, the \storycloze \cite{mostafazadeh_corpus_2016, chaturvedi_ask_2016}, and its generative versions \cite{guan_story_2019}. These tasks might require some common sense reasoning on top of semantic understanding; the fact that they aim to predict the future might require a deeper level of understanding than the previous tasks.

\noindent\textbf{C: What can we infer about the characters?}
 When people read a narrative, they not only grasp the facts explicitly stated in the story, but also make inferences about the actors' mental states, such as their attitudes and desires, as the story unfolds.\footnote{See \citet{bratman_intention_1987} for an account of the Belief-Desire-Intention model of human practical reasoning.} Oftentimes, such an understanding requires inference, either logical or based on common sense reasoning. Such tasks can aim to generate the likely intents and reactions from the actors involved in the events \cite{rashkin_modeling_2018, rashkin_event2mind_2018}, 
or to determine whether a given desire of the protagonist was fulfilled \cite{rahimtoroghi_modelling_2017}.

\noindent\textbf{D: What is the intention of the narrator in sharing their story?} While these prior tasks cover a wide range of angles to narrative understanding, they take an intradiegetic view by focusing on understanding the story itself. We propose another dimension to this line of work by
taking an outside-the-story (extradiegetic) perspective\footnote{Recognizing the importance of these two different perspectives for story understanding, \citet{swanson_empirical_2017} attempted to classify narrative clauses into intradiegetic vs. extradiegetic levels.} and aiming to understand {\it why} the story is shared by the narrator, potentially inferred from {\it how} the narrator decides to construct it. In particular, the task introduced here is to infer the \emph{advice-seeking} intention of the narrator.\footnote{Sharing personal stories can have other goals, such as in therapeutic writing \cite{pennebaker_writing_1997,pennebaker_forming_1999,pennebaker_expressive_2014}.}

We argue that solving this task requires not only the semantic understanding and common sense reasoning involved in prior tasks but also a certain degree of social intuition. To uncover the goals of the narrator, one needs to find cues in the narrative construction---what has been selectively included or emphasized, and what might have been purposefully omitted \cite{labov_transformation_1972}. In fact, such intention-understanding tasks are often included in  ``social intelligence'' tests \cite{conzelmann_new_2013,baumgarten_cue_2015}.

\subsection{Advantages of cloze test formulation} 
\label{sec:formulation}

To evaluate the capacity of NLP systems to solve this task, we consider a binary choice cloze test formulation for two main reasons. First, it allows natural ground-truth labels: often, when people share their personal experiences to seek advice, they add explicit requests for the information they are seeking. After removing these requests from the narratives, we can use them as proxies for the narrators' intentions.  Second, the 
binary choice operationalization
also has the advantage of non-ambiguity in evaluations and ease of comparisons between systems (as opposed to a generation task).

It is worth noting that our dataset is constructed in a way that allows easy modifications into other task formats if so desired. For instance, 
the methodology of identifying a plausible false choice for a given narrative could be applied multiple times to extend the task to a more difficult multiple-choice version. 
Similarly, by ignoring the incorrect question in each instance, our dataset 
can be used as a source for a new generation task, i.e., generating the advice-seeking question from the given narrative.

\section{Task implementation}
\label{sec:method}

For a meaningful implementation of the proposed task, the collection of test instances must conform to several expectations, in terms of both the narratives and their (actual) advice-seeking questions.  
In what follows we outline these desiderata and our method for collecting instances that meet them (Section \ref{sec:procedure}). 

Furthermore, as with any multiple-choice cloze test formulation, the difficulty of each test instance largely depends on how plausible the alternative answers are.  
Yet, finding plausible (but not actually correct) alternatives automatically is challenging.  
Not surprisingly, many of the cloze-style multiple-choice datasets use humans to write these alternatives \cite{mostafazadeh_corpus_2016,xie_large-scale_2018}, 
limiting their scalability.

We tackle this challenge by developing a methodology that exploits both the commonalities in human experiences shared online and the diversity in the types of advice needed for similar situations under different circumstances (Section~\ref{sec:pairing}).

\subsection{Collection of candidate instances}
\label{sec:source}
\label{sec:procedure}

\begin{table*}[ht!]
    \centering
    \begin{tabular}{p{0.18\textwidth}p{0.2\textwidth}p{0.57\textwidth}}
        \toprule
        {\bf Selected topics} & {\bf Question keywords} & {\bf Example questions} \\
         \midrule
        Housing & move live house city apartment roommate  & \parbox[t]{0.57\textwidth}{What is it like living with {\bf roommates}? \\ Should I {\bf move} to the {\bf city}?} \\
        \midrule
        School  & college school class degree study & \parbox[t]{0.57\textwidth}{Should I drop out of {\bf college}? \\ What's the best way for me to {\bf study} for my biology tests?} \\
        \midrule
        Work  & job boss quit work interview employer  & \parbox[t]{0.57\textwidth}{Can I somehow ask to {\bf work} from home?\\ How do I explain during an {\bf interview} why I left a {\bf job}?} \\
        \midrule
        Relationships & girl date text tell guy think crush & \parbox[t]{0.57\textwidth}{Does it sound like this {\bf girl} may like me? \\ How can I think of a better greeting for online {\bf dating}?} \\

        \midrule
        Personal finances  & money car pay rent loan insurance & \parbox[t]{0.57\textwidth}{How do I afford a {\bf car} in my situation? \\ Am I stupid for wanting a student {\bf loan}?} \\
        \midrule
        Family & parent convince let mom dad sister   & \parbox[t]{0.57\textwidth}{How do I {\bf convince} my {\bf parents} to believe me? \\ How can I try and make a better relationship with my {\bf sister}?} \\ 
         \bottomrule
    \end{tabular}
    \caption{Selected narrative topics and example question keywords associated with each topic.}
    \label{tab:advice_properties}
\end{table*}

\xhdr{Narratives desiderata} As a pre-requisite, we need to start from personal narratives containing advice-seeking needs that are explicitly expressed (as questions), and that can be removed to form the cloze test instances.\footnote{An interesting future work avenue could be considering narratives that only have implicit advice-seeking intentions.} 
Ideally, these narratives would cover a broad range of topics, in order to be able to test how well a system can generalize to a diverse range of real-life scenarios, rather than apply only to restricted and artificial settings.

\xhdr{Question desiderata} Not all questions contained within an advice-seeking narrative are suitable for our task.  
Some of the questions might be too general, while others might be rhetorical.  
For instance, {\it Any advice?}  holds no particular connection with the context of the narrative in which it appears. 
To contribute to meaningful test instances, questions need to meet a level of relevance and specificity such that (at least) humans could match them with the narratives from which they are extracted.

\xhdr{Data source} We start from a dataset of over 415,000 advice-seeking posts
 collected from the subreddit r/Advice, which self-defines as ``a place where anyone can seek advice on any subject".\footnote{We start from an existing collection of Reddit posts \cite{tan_all_2015} which we supplement with The Baumgartner Reddit Corpus retrieved via Pushshift API on Nov. 21, 2018.} 
We only use publicly available data and will honor the authors' rights to remove their posts.

\begin{figure}[t!]

          \begin{tabular}{p{7cm}}
          \toprule 
            {\bf Title:} \colorbox{lightgray}{How can I energize myself?} \\
            \hline
            I am generally a person who needs a lot of sleep [...] I don't want to do anything too unhealthy such as drinking 10 cups of coffee, tho I may consider drinking another one. \colorbox{lightgray}{Help? What has worked for you?} \\
            \midrule
        \end{tabular}

\caption{Cloze application to the post from which we obtain the introductory test instance. After filtering out questions that are too general, only the title question remains as a candidate for representing the actual advice-seeking intention of the narrator.}
\label{fig:cloze}
\end{figure}

\xhdr{Applying cloze} For each post, we strip off all questions that appear in any position of the post, including the post title.\footnote{To identify questions, we use the simple heuristic of looking for sentences that end with `?' 
or start with {\it why, how, am, is, are, do, does, did, can, could, should, would}. %
}
We keep the remaining narratives as the cloze texts.\footnote{To ensure that the cloze text can provide sufficient context, yet are not overly verbose, we only consider cloze texts that are 50-300 tokens long. This is a choice we made prior to any experiments, and we do not claim it is the optimal range to set up the task.} 
Figure \ref{fig:cloze} shows how the cloze transformation is applied to the post containing our introductory example.

\xhdr{Selecting ground-truth test answers}\footnote{As it happens, test answers are actually questions.} We select candidate ground-truth answers for the cloze test as the \mbox{?-ending} sentences removed from narratives. In order to keep only well-formed information-seeking questions, we filter the candidate questions by keeping only those that start with interrogatives\footnote{We consider the following set of words as interrogatives: \emph{what, when, why, where, which, 
            who, whom, whose, how, 
            am, is, are, was, were, 
            do, does, did, has, have, had,
            can, could, shall, should, will, would, may, might, must}.} or \emph{any, anyone, help, advice, thoughts}.  To further discard questions that are too general, we compute a simple specificity score $S(q)$ of a question $q$ containing the set of words $\{w_1, w_2, \dots, w_N \}$ as its maximum inverse document frequency (idf):
\[
S(q) = S(\{w_1, w_2, \dots, w_N \}) = \max_{i \in N} \textnormal{idf}(w_i),
\]
and filter out questions for which $S(q)<5$ or questions that have less than 5 words.  At the end of this selection process, from the example post in Figure~\ref{fig:cloze}, \emph{Help?} and \emph{What has worked for you?} are discarded and the title question is kept as the ground-truth answer to this cloze instance.  If multiple questions survive the filtering process, we select one at random.

\xhdr{Diversity evaluation}
To verify that the resulting data has broad topical diversity in both narratives and questions, we perform a two-step clustering analysis.
First, we use singular value decomposition on tf-idf transformed narratives to obtain their vector representations, we then cluster similar narratives using k-means to surface underlying topics.
Next, for each topic, we extract nouns and verbs from the questions attached to each narrative in the topic, and surface common question keywords as those with high document frequency within the topic, correcting for their global document frequency (via subtraction).

To provide a qualitative feel of the diversity of the data, Table \ref{tab:advice_properties} shows a 
 selection of the resulting narrative topics and question keywords, together with example questions (corresponding narratives can be found in the data release).  
We find a wide range of experiences represented in the narratives, from relationships to student life to apartment rentals.
Furthermore, within each narrative topic, there is a variety of question types; for instance, questions related to housing could be about dealing with roommates, paying rent, or choosing a city to live in.

\subsection{Finding alternative test answers}
\label{sec:pairing}

To find plausible alternative answer options for each candidate cloze test instance, one direct approach could be to find questions that are semantically related to the ground-truth question.  However, there are two underlying problems with this approach. First, the task of finding semantically similar questions is itself very challenging \cite{haponchyk_supervised_2018}, given their terseness and lack of context.  
Second, semantic similarity is arguably a different concept from {\it plausibility} with respect to a narrative.  
For example, the two questions in the introductory example are semantically distant, but they are both plausible in the context of the narrative.

Our main intuition in solving this problem is that individuals who are in similar situations tend to have advice-seeking intentions that are related.  For each candidate cloze test narrative instance, we can thus search for a similar narrative first (by exploiting commonalities in experiences people share online) and then select an advice-seeking question from that narrative as the \textit{alternative} answer for the test.

\xhdr{Narrative pairing} 
To operationalize this intuition, we first find pairs of similar narratives based on the cosine similarity of their tf-idf representations.\footnote{We consider both unigrams and bigrams, and set a minimum document frequency of 50. 
We also remove likely duplicates (cosine $>0.8$) and cases for which the similarity between narratives is too low (cosine $< 0.1$). We have also experimented with embedding-based representations to compute cosine similarities from, but they do not seem to produce qualitatively better pairings upon inspection.} A greedy search based on this similarity metric results in a set of pairs of related narratives (\naone, \natwo) with their respective advice-seeking questions (\qnone, \qntwo) identified in the previous step.

\begin{figure}[t!]

  \begin{tabular}{ p{7.5cm} }
  \toprule
{\bf Masked narrative}: I've noticed something, over the past few years I've gained a habit of drinking coffee. The average day is about six cups, but it can exceed that sometimes (8 or so). The only reason I question my habit is cause I'm up at 4AM right now cause I couldn't fall asleep. I honestly have a headache in the morning until I drink a cup of coffee. I'll have some for essentially no reason, I'll just make some out of a urge almost. \\
\midrule
{\bf Q1}: Is it even possible to be addicted to coffee?\\
{\bf Q2}: How can I energize myself?\\
\bottomrule

\end{tabular}
\caption{Alternative cloze test instance corresponding to the introductory example.}

\label{tab:pair}
\end{figure}

\xhdr{Narrative masking}  At this point, the pair of advice-seeking questions could be used with either narrative to form a test instance.  For example, Figure~\ref{tab:pair} shows the other possible cloze instance corresponding to the introductory example if we were to use the other narrative in the narrative pair.  This, however, would arguably be a poor test instance since Q2 is hardly applicable to this other narrative.  More generally, we want to ensure that our choice of which narrative (\nai) to include in the cloze test optimizes the plausibility of the question pair (\qnone, \qntwo).

To achieve this, we compute the similarity between each narrative in the pair and each of the two respective questions,\footnote{To account for the terseness of the questions, we represent both narratives and questions with tf-idf weighted GloVe embeddings \cite{pennington_glove_2014} and compute the cosine similarity between them.} and select the narrative that maximizes the minimum question-narrative similarity.  Formally,
\[
N_i = \argmax_i \textsc{min} \{sim(N_i, qn_1), sim(N_i, qn_2)\}.
\]
Importantly, this selection criterion is purposely symmetric with respect to the two questions in order to avoid introducing any unnatural preference between the two that a classifier (with no access to the masked narrative) could exploit. 

As a final check, we ensure that in each cloze instance the two questions are neither too similar to each other (and thus indistinguishable) nor too dissimilar %
(which may indicate unsatisfactory narrative pairings).
To this end, we discard instances in which the questions have extremely high or low surface similarity according to their InferSent \cite{conneau_supervised_2017} sentence embeddings.\footnote{We set a lower bound of 0.8 and an upper bound of 0.95. We choose this representation because questions are short and thus we anticipate tf-idf representation to be less informative.} 

This process leaves us with a total of 21,865 instances. A detailed account of the number of instances filtered at different stages of the construction process can be found in the Appendix.

\section{Human performance}
\label{sec:eval}

To understand the feasibility of the task, as well as the relative difficulty of the items in the dataset, eight non-author annotators labeled a random sample of 
200 instances.\footnote{See the Appendix for detailed annotation instructions.}
Each annotator is asked to choose first, out of the two candidate questions, which they consider to be {\it more likely} to have been asked by the narrator. Overall, human annotators achieve an accuracy of 90\% (Cohen's $\kappa$ = 0.79),\footnote{We obtained a second round of annotations on a subset of 75 task instances to compute agreement statistics.} showing that humans can indeed recover the advice-seeking intentions of the narrators, and thus validating the feasibility of the task.\footnote{By construction, random accuracy is 50\%.}

We are also interested in understanding the types of skills needed to solve the task.  In particular, we want to estimate the proportion of the task instances that can not be solved by mere factual reasoning.  To this end, we ask humans to identify candidate questions that contain a factual mismatch with the narrative, making them \textbf{E}xplicitly incompatible; 57\% of the annotated instances do not contain any such mismatches in any of the questions.  Similarly, we want to estimate how many instances require common sense expectations about the behavior of the protagonist (within the story).  So we ask annotators to mark questions as being \textbf{I}mplicitly incompatible if they do not contain any factual mismatches, but they are incompatible with what can be inferred implicitly about events and characters in the story.

\begin{table}[ht!]

    \begin{tabular}{l}
        \toprule
        \parbox[t]{0.48\textwidth}{\textbf{Narrative}: I asked a girl that I really like if she would like to get coffee sometime. She said she's really busy but that we'll see. I can't get her off my mind and I spend all day waiting for her to tell me she's free. \\

        \textbf{Explicitly incompatible (E)}:\\
        How to deal with my roommate?\\ 

        \textbf{Implicitly incompatible (I)}:\\
        What to do if I asked a girl out and now regret it?\\

        \textbf{Compatible (C) but unlikely (U)}:\\
        Which coffee place would you recommend?\\

        \textbf{Compatible (C) and likely (L)}:\\
        Would it seem desperate if I asked her again in a more direct way a week later?}\\ 

        \bottomrule
    \end{tabular}
    \caption{Example questions in each plausibility category for an example narrative.}
    \label{tab:category_examples}
\end{table}

\begin{table}[!ht]
    \begin{center}
    
    \begin{tabular}{p{.098\textwidth } p{.035\textwidth} p{.0677\textwidth} p{.065\textwidth} p{.099\textwidth}}

    \toprule
    {\bf Pair type} &  {\bf \textsc{sim}} & {\bf \textsc{ft-lm}} & {\bf \textsc{human}}  & {\bf \% in data} \\
    \midrule
    C + E            & 86\%   & \hspace{0.01in} 88\%   & \hspace{0.01in} 100\%  & \hspace{0.1in} 38\%\\ 
    C + \{C, I\}     & 68\%   & \hspace{0.01in} 74\%   & \hspace{0.01in} 89\%   & \hspace{0.1in} 46\%\\
    C + C            & 66\%   & \hspace{0.01in} 73\%   & \hspace{0.01in} 84\%   & \hspace{0.1in} 32\% \\
    L + \{U, I\}     & 75\%   & \hspace{0.01in} 75\%   & \hspace{0.01in} 100\%  & \hspace{0.1in} 30\% \\
    \textsc{overall} & 76\%   & \hspace{0.01in} 80\%   & \hspace{0.01in} 90\%   & \\
    \bottomrule

    \end{tabular}%

    \caption{Breakdown of performances on selected \textsc{actual} + \textsc{alternative} question pair types. For instance, the pair type C + E corresponds to instances where the \textsc{actual} question asked by the narrator is compatible and the \textsc{alternative} question is explicitly incompatible.}
    \label{tab:breakdown}
\end{center}
\end{table}

\cut{
\begin{table}[!ht]
    \begin{center}
    
    \begin{tabular}{l l l l l}
    \toprule
    {\bf Pair type} &  {\bf \textsc{sim}} & {\bf \textsc{ft-lm}} & {\bf \textsc{human}}  & {\bf \% in data} \\
    \midrule
    C + E             & 86\%  & 88\% & 100\% & 38\%\\ 
    C + \{C, I\}    & 68\%  & 74\% & 89\%  & 46\%\\
    C + C              & 66\%  & 73\% & 84\%  & 32\% \\
    L + \{U, I\}    & 75\%  & 75\% & 100\% & 30\% \\
    \textsc{overall} & 76\% & 80\% & 90\% & \\
    \bottomrule

    \end{tabular}%

    \caption{Breakdown of performances on selected \textsc{actual} + \textsc{alternative} question pair types. For instance, the pair type C + E corresponds to instances where the \textsc{actual} question asked by the narrator is compatible and the \textsc{alternative} question is explicitly incompatible.}
    \label{tab:breakdown}
\end{center}
\end{table}
}

The questions that are neither explicitly nor implicitly incompatible would be labeled as being \textbf{C}ompatible, and as either \textbf{L}ikely or \textbf{U}nlikely to represent the narrators' intentions. Test items in our data forcing a choice between \textbf{C}ompatible questions are expected to be the hardest to solve, as they might require a certain degree of social intuition in addition to factual and common sense reasoning. Table~\ref{tab:category_examples} provides an example narrative and one representative question from each of the above-mentioned categories.\footnote{The example is adapted from our instructions to annotators, which includes further explanations for these categories. See the Appendix for details.}

Table \ref{tab:breakdown} shows a human performance breakdown according to some of the most common types of instances in our data.\footnote{See the Appendix for some representative examples for selected question pair types in our data.}
As expected, instances
involving only compatible questions (C + C) are harder
to solve,\footnote{We also concede that some of the instances in this category may be unsolvable, e.g., when the wrong question fits the narrative just as well.} as they might require some
social intuition, whereas when
explicit contradictions exist (C + E), they are perfectly solvable. We also note that humans can perfectly solve the subset of task instances (L + \{U, ~I\}) that exhibit perceived qualitative differences
    between the actual and the alternative questions, but nevertheless, require more than semantic understanding (and sometimes require social intuition).

\section{Baseline systems performance}
\label{sec:baseline}

We divide our data into a 8,865-2,500 train-test split and have reserved 10,000 instances as a held-out set.\footnote{The set annotated by humans is disjoint.} In Table~\ref{tab:baseline} we report accuracy for the best-performing model on the (never-before-seen) held-out for a simple similarity-based method and for a deep learning method.

\xhdr{Narrative-question similarity} We expect that questions would show greater similarity to narratives they are removed from. We thus establish a narrative-question similarity baseline by considering features based on cosine similarities between narrative and questions, with text represented as tf-idf vectors, tf-idf weighted GloVe embeddings, averaged GloVe embeddings, as well as word overlap between content words, all combined in a logistic regression model.

\xhdr{Finetuned transformer LM} We also use a Finetuned Transformer LM model \cite{radford_improving_2018}, which was shown to perform competitively  on a diverse set of NLP tasks, achieving state-of-the-art results on the \storycloze.\footnote{We fine-tune with our training set on top of the pre-trained transformer language model, using the implementation from \url{https://github.com/huggingface/pytorch-openai-transformer-lm}.} 

\begin{table}[t]
    \begin{center}
    \resizebox{7cm}{!}{%
    \begin{tabular}{l c }
    \toprule
    {\bf Model} &  {\bf Accuracy (held-out)} \\
    \midrule
    \textsc{narrative-qn-sim}  &  73.4\%   \\
    \textsc{finetuned lm}  &  \textbf{78.7\%}  \\
    \bottomrule
    \end{tabular}%
    }
    \caption{Performance of different baselines.} 
     \label{tab:baseline}
\end{center}
\end{table}

\subsection{Error analysis}
\label{sec:error}

\xhdr{Required skills} As shown in Table \ref{tab:breakdown}, systems perform worst on items that do not exhibit any (implicit or explicit) mismatches (C + C), and thus might require some social intuition. Importantly, the largest gap between baseline and human performance (25\%) is on the subset of items that can not be solved based solely on a semantic understanding (L + \{U, I\}). These results underline the need for models that can combine common sense reasoning about the events within the story with an intuition about the intention of the narrator.

\xhdr{Question concreteness} Questions may also differ in how concrete they are. In a preliminary analysis aimed at understanding how this property affects performance, we compare words used in ground-truth questions that the best-performing model predicts correctly with those used in questions that are classified incorrectly. We observe that questions that are predicted correctly have significantly higher average inverse document frequencies (t-test $p<0.01$). Intuitively, these more specific questions may be more concrete in nature, making them easier to connect to the narratives to which they belong. We also find that some common interrogatives have skewed distributions. For instance, questions starting with {\it Is} are less likely to be classified correctly than those starting with {\it How}. 
A cursory manual investigation suggests that this can also be tied by concreteness, with the latter type of questions appearing to be more concrete than the former.

\section{Further related work}
\label{sec:related}

One broad motivation behind our work is to eventually help better support 
personalized informational needs \cite{teevan_characterizing_2007}. This connects to several related lines of work that were not previously discussed.

\xhdr{Query/question intents}
Datasets and models are proposed for understanding user intents behind search queries \cite{radlinski_inferring_2010,fariha_squid_2018}, or even more generally, user questions \cite{haponchyk_supervised_2018}. To complement this line of work that looks at user intents behind the explicit request, our task aims to uncover user intents when they are implied in personal narratives (without access to the explicit question). 

\xhdr{Conversational search/QA} One way to better satisfy user intents is by making such processes collaborative \cite{morris_searchtogether_2007,morris_collaborative_2013}, or conversational \cite{radlinski_theoretical_2017}. Conversational QA datasets \cite{choi_quac_2018,reddy_coqa_2019} have been introduced to help develop systems with such capability. 

\xhdr{Social QA} Some questions posed by users are inherently more social in nature, and require more nuanced contextual understanding \cite{harabagiu_questions_2008}. The social nature may affect how people ask questions \cite{dahiya_discovering_2016,rao_learning_2018}, and pose challenges for identifying appropriate answers \cite{shtok_learning_2012,zhang_detecting_2017}.

\section{Discussion}
\label{sec:discussion}

In this work, we introduce the new task of inferring advice-seeking intentions from personal narratives, a methodology for creating appropriate test instances for this task and the \asq dataset.
This task complements existing (intradiegetic) narrative understanding tasks by focusing on extradiegetic aspects of the narrative: in order to understand ``Why is the narrator sharing this?'', we often need to apply a certain degree of common sense and social intuition.

From a practical perspective, this extradiegetic capability is a prerequisite to properly address 
personalized information needs that are constrained by personal circumstances described as free-form personal stories.  
Currently, to address these types of information needs, people seek (or even hire) other individuals with relevant experience or expertise.  
As with conversational search \cite{radlinski_theoretical_2017}, we can envision systems that can more directly address complex information needs by better understanding the circumstances and intentions of the user. 

Our analysis of the human 
and baseline performance on different types of test instances points to interesting avenues for future work, both in terms of designing better-performing systems and in terms of 
constructing better test data. 
We envision that (intradiegetic) narrative understanding could help identify the components of the narrative that are most relevant to the advice-seeking goal. For example, identifying the narrator's intentions and desires within the story \cite{rashkin_event2mind_2018}, and whether these desires are fulfilled \cite{rahimtoroghi_modelling_2017} could help focus the attention of the model, especially when dealing with less concrete questions.
   Furthermore, a better representation of the structure of the narrative \cite{ouyang_towards_2014}, in terms of discourse acts \cite{elson_modeling_2012}
    and sentiment flow \cite{ouyang_modeling_2015}, could also help distinguish between spurious and essential circumstances of the narratives.

In terms of improving the task itself and the methodology for creating testing instances that better approximate the inferential task, we note a few possible directions. Firstly, better narrative modeling could lead to higher quality matching.  Similarly, better representation of the questions can help select more appropriate candidate options (e.g., currently 6\% of the questions are deemed by the annotators to be too general). In addition, the generative version of the task, when appropriately evaluated, could be a closer operationalization for intention inference, and also offer more potential for practical uses.  

Finally, future work could expand on our methodology to formulate other more general tasks aiming to understand the reasons why a person is sharing a personal story. While we have focused on narratives shared with the intention of seeking advice, people may also share stories to express emotions, to entertain or educate others. A better understanding of these different (explicit or implicit) intentions could lead to more personalized and empathetic human-computer interaction.

\vspace{0.2in}

\xhdr{Acknowledgments} The authors thank Tom Davidson, Tsung-Yu Hou, Qian Huang, Hajin Lim, Laure Thompson, Andrew Wang, Xiaozhi Wang and Justine Zhang for helping with the annotations. We are grateful to Thorsten Joachims, Avery Quinn Smith and Todd Cullen for helping us 
when our server crashed on the day of the deadline while testing the model on the held-out set, to Lillian Lee, Andrew Wang, Justine Zhang and the anonymous reviewers for their helpful comments, and to Fernando Pereira for the early discussions that inspired this research direction. This work is supported in part by NSF CAREER award IIS-1750615 and NSF Grant SES-1741441.

\bibliographystyle{acl_natbib}

\bibliography{narratives-naacl-autoupdate-LF,narratives-naacl-autoupdate-C,narratives-naacl-autoupdate-JPC}

\vspace{0.2in}

\section*{Appendix}
\appendix
\renewcommand{\thesection}{A\arabic{section}}
\renewcommand{\thetable}{A\arabic{table}}

\maketitle

\section{Instructions to human annotators}

As described in Section 4 of the main paper, we obtained human annotations on a small subset of our data for validation purposes.
Annotators were shown instructions which included the definitions for different question-narrative plausibility types, together with two examples to help further clarify the task and the definitions. The exact instructions are reproduced in Table \ref{fig:instructions}, while the examples provided are shown in Table \ref{tab:appendix_instruction_examples}.

\begin{table}[t]
\centering
    \begin{tabular}{lll}
    \toprule
        {\bf Question type} & {\bf \parbox[b]{.07\textwidth}{\% in\\ actual}} & {\bf \parbox[b]{.07\textwidth}{\% in\\ altern.}} \\
     
     \midrule
         \textbf{C}ompatible and \textbf{L}ikely  & 81\% & 15\% \\
         \textbf{C}ompatible but \textbf{U}nlikely &  10\% & 21\% \\
         Incompatible (\textbf{I}mplicit) & 5\% & 16\% \\
         Incompatible (\textbf{E}xplicit) & 2\% & 41\% \\
         Very \textbf{G}eneral &  3\% & 8\% \\
         \bottomrule

    \end{tabular}
    \caption{Data distribution estimated from the human annotated subset. For more than half of the cases, the wrong answer (i.e., the alternative question) could not be simply discarded based on factual mismatches, and the task instance would require additional common sense or social intuition to solve.}
    \label{tab:distribution}
\end{table}

\begin{table}[h!]    
    \begin{tabular}{l l}
    \toprule
    \# of unique post ids &  415,693 \\
    \# of posts with narrative bodies & 339,815 \\
    \# of narratives with questions & 262,721 \\
    \# of narratives after filtering for length & 151,418 \\
    \# of narratives with specific questions & 89,527 \\ 
    \# of narratives paired & 43,730 \\
    \bottomrule

    \end{tabular}%

    \caption{Counts for narrative instances at different stages of dataset construction.}
    \label{tab:filtering}
\end{table}

\begin{table*}[ht!]

\begin{tabular}{p{.95\textwidth}}

    \toprule

    You will be presented with one narrative and two advice-seeking questions (qn1 and qn2). \\ \\

    Firstly, you will need to indicate which of these questions is \textbf{more likely} to be asked by the narrator in the context of the narrative (Column D, in yellow). Use the dropdown menu to select the more likely question among the two.  (You must pick one.) \\ \\

    In addition, for \textbf{each} question separately, you will need to provide a rating on how plausible the question is in the context of the narrative by choosing one of the following options  (dropdown menu in Columns E and F, in green):

    \begin{enumerate}
    \item \textbf{Very general}: i.e., this question could follow most narratives, and it's not in any way specific to this narrative.
    \item \textbf{Compatible and likely}: i.e., the question follows naturally from this narrative.
    \item \textbf{Compatible but unlikely}: i.e., while there is no direct contradiction (either explicit or implicit) with the narrative, it seems unlikely that the narrator's intention was to ask this question. 
    \item \textbf{Incompatible (explicit)}: i.e., the question is incompatible due to clear factual mismatches with the information explicitly contained in the narrative, or it is completely irrelevant.
    \item \textbf{Incompatible (implicit)}: i.e., the question is incompatible due to mismatches with something that you can indirectly infer from the narrative.
    \end{enumerate}
    
    You should judge each question \textbf{separately} when selecting a category.  It is possible for both questions to fall into the same category. \\ \\

    Now you need to read the example narratives, questions and explanations in the adjacent cells (B2 and C2) to get a feel of each of the categories, after which you could proceed to the sub-sheet \textbf{Items to annotate (see tab at the bottom of this page)} to complete the annotation task. \\ \\ 

    Optionally, you can also provide comments for each item (scroll to the right to see the comment column): Did you find an item particularly challenging or interesting?  Was one of the questions not really asking for an advice?  Do share your thoughts with us. \\ 

    \bottomrule

 \end{tabular}
\caption{Instruction text shown to the annotators.}
\label{fig:instructions}
\end{table*}

\section{Distribution of plausibility categories}

Table \ref{tab:distribution} shows the distribution of plausibility categories given out by our human annotators, for both the actual questions which belong to the given narrative (Column 2) and the paired alternative questions (Column 3).

\section{Further processing details}

Table \ref{tab:filtering} provides the number of instances remaining at each of the processing steps. After masking one narrative from each narrative pair, we have a total of 21,865 narratives, each successfully paired with two plausible candidate questions.

\begin{table*}[ht!]
    \centering
    \begin{tabular}{ll}
        \toprule
        {\bf Example 1} & {\bf Example 2} \\
        \midrule
        \parbox[t]{0.48\textwidth}{Narrative: ``I am a freshman in college and I have a group of friends that I have been hanging out with for the past couple months. I feel like we have a good time when we hang out, but a lot of the time, the rest of the group will go out and do stuff together, but I won't be included.'' \\ \\ 
        Example questions in each category [and explanations where appropriate]:\\ \\ 
        a) \textbf{Very general}:\\  
        Any advice?\\ \\ 
        b) \textbf{Compatible and likely}:\\ 
        Can I ask to be included?\\ \\ 
        c) \textbf{Compatible but unlikely}:\\ 
        How to make new friends in college? \\ 
        $[$explanation: it is more likely that the narrator is trying to be more included in the current group of friends, rather than giving up entirely on them and look for replacement.$]$\\ \\ 
        d) \textbf{Incompatible (explicit)}:\\ 
        Any advice for finding new friends for a senior in college?\\ 
        $[$Explanation: The narrator is a freshman in college, not a senior. This constitutes a clear factual mismatch between the question and the narrative.$]$\\ \\ 
        e) \textbf{Incompatible (implicit)}:\\ 
        What are some excuses to not hang out with them?\\ 
        $[$Explanation: We can imply from the narrative that the narrator wants to hang out with the group.  This is incompatible with a question asking how NOT to do that.$]$} & \parbox[t]{.48\textwidth}{Narrative: ``I asked a girl that I really like if she would like to get coffee sometime. She said she's really busy but that we'll see. I can't get her off my mind and I spend all day waiting for her to tell me she's free.''\\ \\
        Example questions in each category [and explanations where appropriate]:\\ \\
        a) \textbf{Very general}:\\
        What should I do?\\ \\
        b) \textbf{Compatible and likely}:\\
        Would it seem desperate if I asked her again in a more direct way a week later ?\\ \\
        c) \textbf{Compatible but unlikely}:\\
        Which coffee place would you recommend? \\
        $[$Explanation: While the narrator is trying to invite the girl for coffee, the main concern seems to be whether the attempt would be successful rather the choices between coffee places.$]$\\ \\
        d) \textbf{Incompatible (explicit)}:\\
        How to deal with my roommate?\\
        $[$Explanation: This question is completely irrelevant to the narrative (no roommate is mentioned).$]$\\ \\
        e) \textbf{Incompatible (implicit)}:\\
        What to do if I asked a girl out and now regret it?\\
        $[$Explanation: We can infer that the narrator is looking forward to the potential date, which contradicts with the feeling of regret in the question.$]$} \\
        \bottomrule
    \end{tabular}
    \caption{Example narratives and questions shown to the annotators.}
    \label{tab:appendix_instruction_examples}
\end{table*}

\section{Examples of pair types}

In our error analysis, we find that the performance of both our baselines, as well as that of our human annotators, vary depending on question pair type, where pair type is defined as the human-judged plausibility of the ground-truth question and that of the alternative question. To give a better sense of what these pair types look like in practice, Table~\ref{tab:pair_examples} shows example instances for a few selected pair types.

\begin{table*}[ht!]
    \centering
    \begin{tabular}{p{.06\textwidth}p{.55\textwidth}p{.16\textwidth}p{.16\textwidth}}
        \toprule
        {\bf Pair type} & {\bf Narrative} & \parbox[t]{0.1\textwidth}{\bf Actual question} & \parbox[t]{0.1\textwidth}{\bf Alternative question} \\
        \midrule
        L + L & Hey everyone I have a bit of a dilemma. It's the first week of school and I am talking three advanced classes, AP world history II, English honors II and Chemistry honors. I am pretty sure that I can handle it but; I am falling behind in chemistry honors and it is the first week. I don't have the mathematical background as the other students. They have taken physics and geometry. I am in a special Algebra class which means I am a year behind in math and science. & Should I drop chemistry honors? & How much do honors courses matter? \\
        L + U & My college roommate/one of my best friends is getting married Saturday. I'm a groomsman, as is our third roommate. Our third roommate gave he and his betrothed their wedding gifts early today: an Xbox One and a crystal decorative bowl from Tiffany. I'm an assistant manager at a sporting goods store making \$8.50 an hour, and between rent, utilities, groceries, gas, and my student loan payments, I usually either barely break even every month or have to borrow money from my parents until my next paycheck. I've checked their registry, and even the less expensive gifts are outside what I can afford (\$30 can make or break me right now). & What to do about a wedding gift if I'm broke? & Where can I buy food that's cheap, and it'll last me until then? \\
        C + I & I just recently switched schools this school year. I'm pretty okay with how it's going so far academics wise but I have no idea how to put myself out there. Everyone has seemed to have made friend groups already or they already know everyone from previous years. I used to be in a private school so no one really knows me from this school except for my close friends that I've known for a long time. & Is there any way that I could gain any popularity before it's too late? & Is it too early to tell if I want to drop out? \\
        L + E & So there is a dream job which is PERFECT for me and of course I really want it. I called the employer last week and she said she was going to call candidates for interviews that week. Then I called this week and she said she was going to call for interviews this week. And please, no advice telling me 'don't call'. I have nothing to lose, so I'm going to call, I would just really appreciate some advice as to how to ask for an interview appropriately - Thank you all! & How do I call an employer asking for an interview? & When I went for the interview she did seem busy so maybe she was too busy to call?
 \\
        L + G & I'm in a relationship with an amazing girl and feel very happy with her. Recently though I've been having intrusive thoughts about her ex-boyfriends (her having sex with them, etc) which are leading to feelings of jealousy and it's really disrupting my ability to enjoy my time with her. & How to deal with feelings of jealousy? & Is this ``normal'' -- in the sense of, do other people experience this? \\
        \bottomrule
    \end{tabular}
    \caption{Example task instances for different \textsc{actual} + \textsc{alternative} question pair types.}
    \label{tab:pair_examples}
\end{table*}

\end{document}